\documentclass[11pt,a4paper]{article}
\usepackage[hyperref]{acl2018}
\usepackage{times}
\usepackage{latexsym}
\usepackage{graphicx}
\usepackage{tabularx}
\usepackage{multirow}
\usepackage{soul}
\usepackage{color}
\usepackage{graphicx}
\usepackage{tabularx}
\usepackage{multirow}
\usepackage{soul}
\usepackage{color}
\usepackage{url}
\usepackage[ruled]{algorithm2e}

\aclfinalcopy 
\def\aclpaperid{38} 

\title{\textbf{\textsc{NCRF++}: An Open-source Neural Sequence Labeling Toolkit}}

\author{Jie Yang \and Yue Zhang \\
  Singapore University of Technology and Design\\
  {\tt jie\_yang@mymail.sutd.edu.sg} \\
  {\tt yue\_zhang@sutd.edu.sg} \\
  }

\date{}

\begin{document}
\maketitle
\begin{abstract}
  This paper describes NCRF++, a toolkit for neural sequence labeling. NCRF++ is designed for quick implementation of different neural sequence labeling models with a CRF inference layer. It provides users with an inference for building the custom model structure through configuration file with flexible neural feature design and utilization. Built on PyTorch\footnote{\url{http://pytorch.org/}}, the core operations are calculated in batch, making the toolkit efficient with the acceleration of GPU. It also includes the implementations of most state-of-the-art neural sequence labeling models such as LSTM-CRF, facilitating reproducing and refinement on those methods. 
\end{abstract}

\section{Introduction}

Sequence labeling is one of the most fundamental NLP models, which is used for many tasks such as named entity recognition (NER), chunking, word segmentation and part-of-speech (POS) tagging. It has been traditionally investigated using statistical approaches \cite{lafferty2001conditional,ratinov2009design}, where conditional random fields (CRF) \cite{lafferty2001conditional} has been proven as an effective framework, by taking discrete features as the representation of input sequence \cite{sha2003shallow,keerthi2007crf}. 

With the advances of deep learning, neural sequence labeling models have achieved state-of-the-art for many tasks \cite{ling2015finding,ma2016end,peters2017semi}. Features are extracted automatically through network structures including long short-term memory (LSTM) \cite{hochreiter1997long} and convolution neural network (CNN) \cite{lecun1989backpropagation}, with distributed word representations. Similar to discrete models, a CRF layer is used in many state-of-the-art neural sequence labeling models for capturing label dependencies \cite{collobert2011natural,lample2016neural,peters2017semi}. 

\begin{figure}[!t]
\scalebox{0.8}{
\begin{tabular}{l}
\texttt{\#\#NetworkConfiguration\#\#}\\
\texttt{use\_crf=True}\\
\texttt{word\_seq\_feature=LSTM}\\
\texttt{word\_seq\_layer=1}\\
\texttt{char\_seq\_feature=CNN}\\
\texttt{feature=[POS] emb\_dir=None emb\_size=10}\\
\texttt{feature=[Cap] emb\_dir=\%(cap\_emb\_dir)}\\
\texttt{\#\#Hyperparameters\#\#}\\
\texttt{...}\\
\end{tabular}
}
\caption{Configuration file segment} 
  \label{fig:config} 
\end{figure}

\begin{figure*}[!t]
  \centering 
    \includegraphics[width=6in]{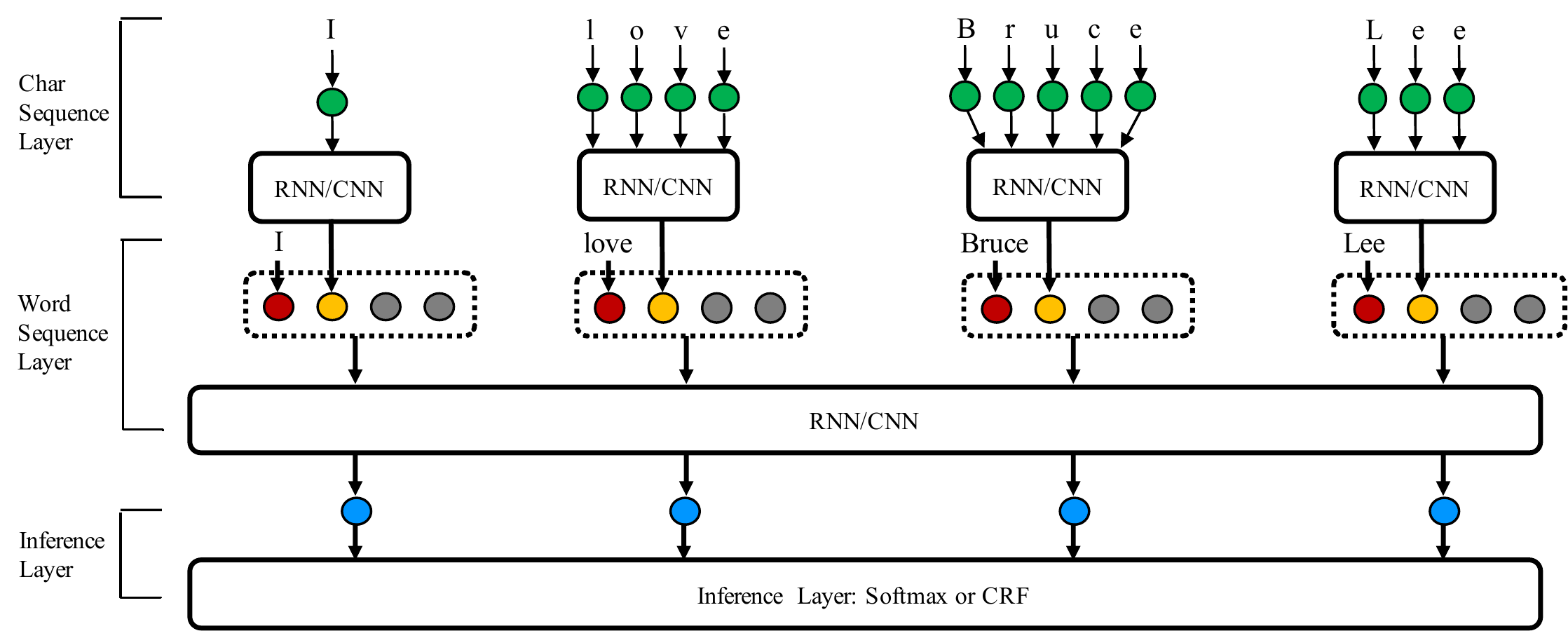}
  \caption{\textsc{NCRF++} for sentence ``I love Bruce Lee''. Green, red, yellow and blue circles represent character embeddings, word embeddings, character sequence representations and word sequence representations, respectively. The grey circles represent the embeddings of handcrafted features.} 
  \label{fig:architecture} 
\end{figure*}

There exist several open-source statistical CRF sequence labeling toolkits, such as CRF++\footnote{\url{https://taku910.github.io/crfpp/}}, CRFSuite \cite{okazaki2007crfsuite} and FlexCRFs \cite{phan2004flexcrfs}, which provide users with flexible means of feature extraction, various training settings and decoding formats, facilitating quick implementation and extension on state-of-the-art models. On the other hand, there is limited choice for neural sequence labeling toolkits. Although many authors released their code along with their sequence labeling papers \cite{lample2016neural,ma2016end,liu2017empower}, the implementations are mostly focused on specific model structures and specific tasks. Modifying or extending can need enormous coding.

In this paper, we present Neural CRF++ (NCRF++)\footnote{Code is available at \url{https://github.com/jiesutd/NCRFpp}.}, a neural sequence labeling toolkit based on PyTorch, which is designed for solving general sequence labeling tasks with effective and efficient neural models. It can be regarded as the neural version of CRF++, with both take the CoNLL data format as input and can add handcrafted features to CRF framework conveniently. We take the layer\-wise implementation, which includes character sequence layer, word sequence layer and inference layer. NCRF++ is:

\noindent \textbullet $\,$ \textbf{Fully configurable}: users can design their neural models only through a configuration file without any code work. Figure \ref{fig:config} shows a segment of the configuration file. It builds a LSTM-CRF framework with CNN to encode character sequence (the same structure as \newcite{ma2016end}), plus \texttt{POS} and \texttt{Cap} features, within 10 lines. This demonstrates the convenience of designing neural models using NCRF++.

\noindent \textbullet $\,$ \textbf{Flexible with features}: human-defined features have been proved useful in neural sequence labeling \cite{collobert2011natural,chiu2015named}. Similar to the statistical toolkits, NCRF++ supports user-defined features but using distributed representations through lookup tables, which can be initialized randomly or from external pretrained embeddings (embedding directory: \texttt{emb\_dir} in Figure \ref{fig:config}). In addition, NCRF++ integrates several state-of-the-art automatic feature extractors, such as CNN and LSTM for character sequences, leading easy reproduction of many recent work \cite{lample2016neural,chiu2015named,ma2016end}.

\noindent \textbullet $\,$ \textbf{Effective and efficient}: we reimplement several state-of-the-art neural models \cite{lample2016neural,ma2016end} using NCRF++. Experiments show models built in NCRF++ give comparable performance with reported results in the literature. Besides, NCRF++ is implemented using batch calculation, which can be accelerated using GPU. Our experiments demonstrate that NCRF++ as an effective and efficient toolkit.

\noindent \textbullet $\,$ \textbf{Function enriched}: NCRF++ extends the Viterbi algorithm \cite{viterbi1967error} to enable decoding $n$ best sequence labels with their probabilities.

Taking NER, Chunking and POS tagging as typical examples, we investigate the performance of models built in NCRF++, the influence of human-defined and automatic features, the performance of \textit{nbest} decoding and the running speed with the batch size. Detail results are shown in Section \ref{sec:eva}.

\section{NCRF++ Architecture}
The framework of NCRF++ is shown in Figure \ref{fig:architecture}. NCRF++ is designed with three layers: a character sequence layer; a word sequence layer and inference layer. For each input word sequence, words are represented with word embeddings. The character sequence layer can be used to automatically extract word level features by encoding the character sequence within the word. Arbitrary handcrafted features such as capitalization \texttt{[Cap]}, POS tag \texttt{[POS]}, prefixes \texttt{[Pre]} and suffixes \texttt{[Suf]} are also supported by NCRF++. Word representations are the concatenation of word embeddings (red circles), character sequence encoding hidden vector (yellow circles) and handcrafted neural features (grey circles). Then the word sequence layer takes the word representations as input and extracts the sentence level features, which are fed into the inference layer to assign a label to each word. When building the network, users only need to edit the configuration file to configure the model structure, training settings and hyperparameters. We use layer-wised encapsulation in our implementation. Users can extend NCRF++ by defining their own structure in any layer and integrate it into NCRF++ easily.

\subsection{Layer Units} \label{ssec:layers}
\subsubsection{Character Sequence Layer}
The character sequence layer integrates several typical neural encoders for character sequence information, such as RNN and CNN. It is easy to select our existing encoder through the configuration file (by setting \texttt{char\_seq\_feature} in Figure \ref{fig:config}). Characters are represented by character embeddings (green circles in Figure \ref{fig:architecture}), which serve as the input of character sequence layer.

\noindent \textbullet $\,$ \textbf{Character RNN} and its variants Gated Recurrent Unit (GRU) and LSTM are supported by NCRF++. The character sequence layer uses bidirectional RNN to capture the left-to-right and right-to-left sequence information, and concatenates the final hidden states of two RNNs as the encoder of the input character sequence.

\noindent \textbullet $\,$  \textbf{Character CNN} takes a sliding window to capture local features, and then uses a \textit{max-pooling} for aggregated encoding of the character sequence.

\subsubsection{Word Sequence Layer}
Similar to the character sequence layer, NCRF++ supports both RNN and CNN as the word sequence feature extractor. The selection can be configurated through \texttt{word\_seq\_feature} in Figure \ref{fig:config}. The input of the word sequence layer is a word representation, which may include word embeddings, character sequence representations and handcrafted neural features (the combination depends on the configuration file). The word sequence layer can be stacked, building a deeper feature extractor.

\noindent \textbullet $\,$ \textbf{Word RNN} together with GRU and LSTM are available in NCRF++, which are popular structures in the recent literature \cite{huang2015bidirectional,lample2016neural,ma2016end,yang2017transfer}. Bidirectional RNNs are supported to capture the left and right contexted information of each word. The hidden vectors for both directions on each word are concatenated to represent the corresponding word.

\noindent \textbullet $\,$  \textbf{Word CNN} utilizes the same sliding window as character CNN, while a nonlinear function \cite{glorot2011deep} is attached with the extracted features. Batch normalization \cite{ioffe2015batch} and dropout \cite{srivastava2014dropout} are also supported to follow the features.

\subsubsection{Inference Layer}
The inference layer takes the extracted word sequence representations as features and assigns labels to the word sequence. NCRF++ supports both softmax and CRF as the output layer. A linear layer firstly maps the input sequence representations to label vocabulary size scores, which are used to either model the label probabilities of each word through simple softmax or calculate the label score of the whole sequence.

\noindent \textbullet $\,$  \textbf{Softmax} maps the label scores into a probability space. Due to the support of parallel decoding, softmax is much more efficient than CRF and works well on some sequence labeling tasks \cite{ling2015finding}. In the training process, various loss functions such as negative likelihood loss, cross entropy loss are supported.

\noindent \textbullet $\,$  \textbf{CRF} captures label dependencies by adding transition scores between neighboring labels. NCRF++ supports CRF trained with the sentence-level maximum log-likelihood loss. During the decoding process, the Viterbi algorithm is used to search the label sequence with the highest probability. In addition, NCRF++ extends the decoding algorithm with the support of \textit{nbest} output.

\subsection{User Interface}
NCRF++ provides users with abundant network configuration interfaces, including the network structure, input and output directory setting, training settings and hyperparameters. By editing a configuration file, users can build most state-of-the-art neural sequence labeling models. On the other hand, all the layers above are designed as ``plug-in'' modules, where user-defined layer can be integrated seamlessly.

\subsubsection{Configuration}
\noindent \textbullet $\,$ \textbf{Networks} can be configurated in the three layers as described in Section \ref{ssec:layers}. It controls the choice of neural structures in character and word levels with \texttt{char\_seq\_feature} and \texttt{word\_seq\_feature}, respectively. The inference layer is set by \texttt{use\_crf}. It also defines the usage of handcrafted features and their properties in \texttt{feature}. 

\noindent \textbullet $\,$ \textbf{I/O} is the input and output file directory configuration. It includes \texttt{training\_dir, dev\_dir, test\_dir, raw\_dir}, pretrained character or word embedding (\texttt{char\_emb\_dim} or \texttt{word\_emb\_dim}), and decode file directory (\texttt{decode\_dir}).

\noindent \textbullet \, \textbf{Training} includes the loss function (\texttt{loss\_function}), optimizer (\texttt{optimizer})\footnote{Currently NCRF++ supports five optimizers: \texttt{SGD/AdaGrad/AdaDelta/RMSProp/Adam}.} shuffle training instances \texttt{train\_shuffle} and average batch loss \texttt{ave\_batch\_loss}.

\noindent \textbullet $\,$ \textbf{Hyperparameter} includes most of the parameters in the networks and training such as learning rate (\texttt{lr}) and its decay (\texttt{lr\_decay}), hidden layer size of word and character (\texttt{hidden\_dim} and \texttt{char\_hidden\_dim}), \textit{nbest} size (\texttt{nbest}), batch size (\texttt{batch\_size}), dropout (\texttt{dropout}), etc. Note that the embedding size of each handcrafted feature is configured in the networks configuration (\texttt{feature=[POS] emb\_dir=None emb\_size=10} in Figure \ref{fig:config}).

\subsubsection{Extension}
Users can write their own custom modules on all three layers, and user-defined layers can be integrated into the system easily. For example, if a user wants to define a custom character sequence layer with a specific neural structure, he/she only needs to implement the part between input character sequence indexes to sequence representations. All the other networks structures can be used and controlled through the configuration file. A \texttt{README} file is given on this.

\begin{table}[!tp]
\begin{center}
\resizebox{\columnwidth}{!}{%
\begin{tabular}{|l|l|l|l|}
\hline 
\multirow{2}*{\textbf{Models}} &\multicolumn{1}{|l|}{\textbf{NER}} &\multicolumn{1}{|l|}{\textbf{chunking}} &\multicolumn{1}{|l|}{\textbf{POS}}\\
\cline{2-4}
 &\textbf{F1-value} &\textbf{F1-value} &\textbf{Acc}  \\ 
\hline
Nochar+WCNN+CRF &88.90&94.23&96.99\\ 
\hline
CLSTM+WCNN+CRF &90.70&94.76&97.38\\ 
\hline
CCNN+WCNN+CRF &90.43&94.77&97.33\\ 
\hline
\hline
Nochar+WLSTM+CRF &89.45&94.49&97.20 \\ 
\hline
CLSTM+WLSTM+CRF &91.20&95.00&97.49\\ 
\hline
CCNN+WLSTM+CRF &\textbf{91.35}&\textbf{95.06}&97.46\\ 
\hline
\hline
\newcite{lample2016neural} &90.94&--&97.51\\ 
\newcite{ma2016end} &91.21&--&\textbf{97.55}\\ 
\newcite{yang2017transfer}&91.20&94.66&\textbf{97.55}\\ 
\newcite{peters2017semi}&90.87&95.00&--\\
\hline
\end{tabular}
}
\end{center}
\caption{Results on three benchmarks.}
\label{tab:allresult}
\end{table}

\section{Evaluation} \label{sec:eva}
\subsection{Settings}
To evaluate the performance of our toolkit, we conduct the experiments on several datasets. For NER task, CoNLL 2003 data \cite{tjong2003introduction} with the standard split is used. For the chunking task, we perform experiments on CoNLL 2000 shared task \cite{tjong2000introduction}, data split is following \newcite{reimers2017reporting}. For POS tagging, we use the same data and split with \newcite{ma2016end}. We test different combinations of character representations and word sequence representations on these three benchmarks. Hyperparameters are mostly following \newcite{ma2016end} and almost keep the same in all these experiments\footnote{We use a smaller learning rate (0.005) on CNN based word sequence representation.}. Standard SGD with a decaying learning rate is used as the optimizer.

\begin{table}[!tp]
\begin{center}
\resizebox{\columnwidth}{!}{%
\begin{tabular}{|l|l|c|c|c|}
\hline 
\multicolumn{2}{|c|}{\textbf{Features}} &\textbf{P} &\textbf{R} &\textbf{F}\\
\hline

Baseline&WLSTM+CRF &80.44&87.88&89.15\\ 
\hline
\multirow{3}*{Human Feature}
&$\;\;\;\;\;$+POS &90.61&89.28&89.94\\
&$\;\;\;\;\;$+Cap &90.74&90.43&90.58\\
&$\;\;\;\;\;$+POS+Cap &90.92&90.27&90.59\\
\hline 
\multirow{2}*{Auto Feature}
&$\;\;\;\;\;$+CLSTM &91.22&91.17&91.20\\
&$\;\;\;\;\;$+CCNN &91.66&91.04&91.35\\
\hline
\end{tabular}
}
\end{center}
\caption{Results using different features.}
\label{tab:feature}
\end{table}

\subsection{Results}
Table \ref{tab:allresult} shows the results of six CRF-based models with different character sequence and word sequence representations on three benchmarks. State-of-the-art results are also listed. In this table, ``Nochar'' suggests a model without character sequence information. ``CLSTM'' and ``CCNN'' represent models using LSTM and CNN to encode character sequence, respectively. Similarly, ``WLSTM'' and ``WCNN'' indicate that the model uses LSTM and CNN to represent word sequence, respectively.

As shown in Table \ref{tab:allresult}, ``WCNN'' based models consistently underperform the ``WLSTM'' based models, showing the advantages of LSTM on capturing global features. Character information can improve model performance significantly, while using LSTM or CNN give similar improvement. Most of state-of-the-art models utilize the framework of word LSTMCRF with character LSTM or CNN features (correspond to ``CLSTM+WLSTM+CRF'' and ``CCNN+WLSTM+CRF'' of our models) \cite{lample2016neural,ma2016end,yang2017transfer,peters2017semi}. Our implementations can achieve comparable results, with better NER and chunking performances and slightly lower POS tagging accuracy. Note that we use almost the same hyperparameters across all the experiments to achieve the results, which demonstrates the robustness of our implementation. The full experimental
results and analysis are published in \newcite{yang2018design}.

\subsection{Influence of Features}
We also investigate the influence of different features on system performance. Table \ref{tab:feature} shows the results on the NER task. POS tag and capital indicator are two common features on NER tasks \cite{collobert2011natural,huang2015bidirectional,strubell2017fast}. In our implementation, each POS tag or capital indicator feature is mapped as 10-dimension feature embeddings through randomly initialized feature lookup table \footnote{\texttt{feature=[POS] emb\_dir=None emb\_size=10}\\ \textcolor{white}{aaaa}\texttt{feature=[Cap] emb\_dir=None emb\_size=10}}. The feature embeddings are concatenated with the word embeddings as the representation of the corresponding word. Results show that both human features \texttt{[POS]} and \texttt{[Cap]} can contribute the NER system, this is consistent with previous observations \cite{collobert2011natural,chiu2015named}. By utilizing LSTM or CNN to encode character sequence automatically, the system can achieve better performance on NER task.

\begin{figure}[!t]
  \centering 
    \includegraphics[width=3in]{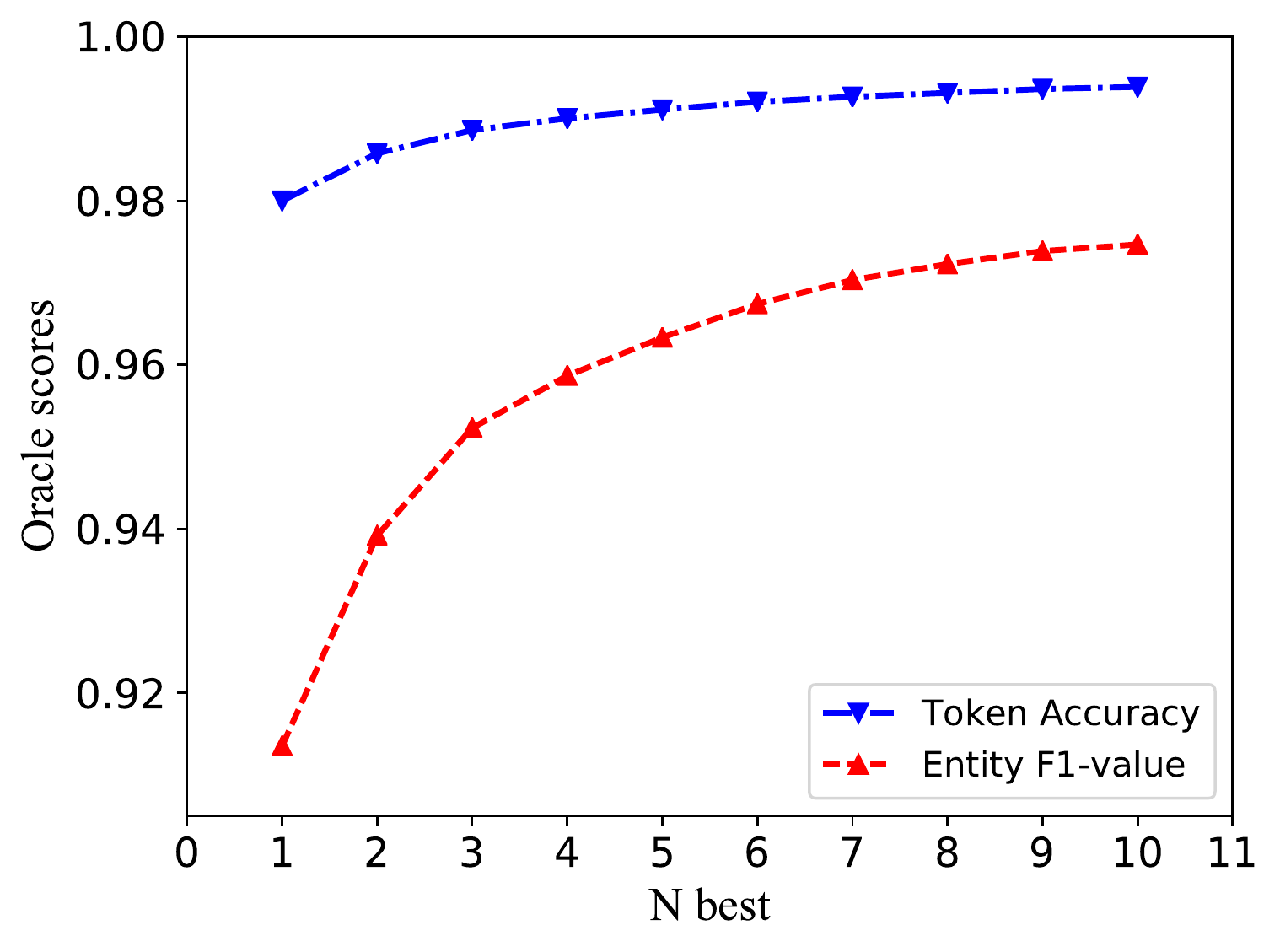}
  \caption{Oracle performance with nbest.} 
  \label{fig:nbest} 
\end{figure}

\subsection{N best Decoding}
We investigate \textit{nbest} Viterbi decoding on NER dataset through the best model ``CCNN+WLSTM+CRF''. Figure \ref{fig:nbest} shows the oracle entity F1-values and token accuracies with different \textit{nbest} sizes. The oracle F1-value rises significantly with the increasement of \textit{nbest} size, reaching 97.47\% at $n=10$ from the baseline of 91.35\%. The token level accuracy increases from 98.00\% to 99.39\% in \textit{10-best}. Results show that the \textit{nbest} outputs include the gold entities and labels in a large coverage, which greatly enlarges the performance of successor tasks.

\subsection{Speed with Batch Size}
\begin{figure}[!t]
  \centering 
    \includegraphics[width=3in]{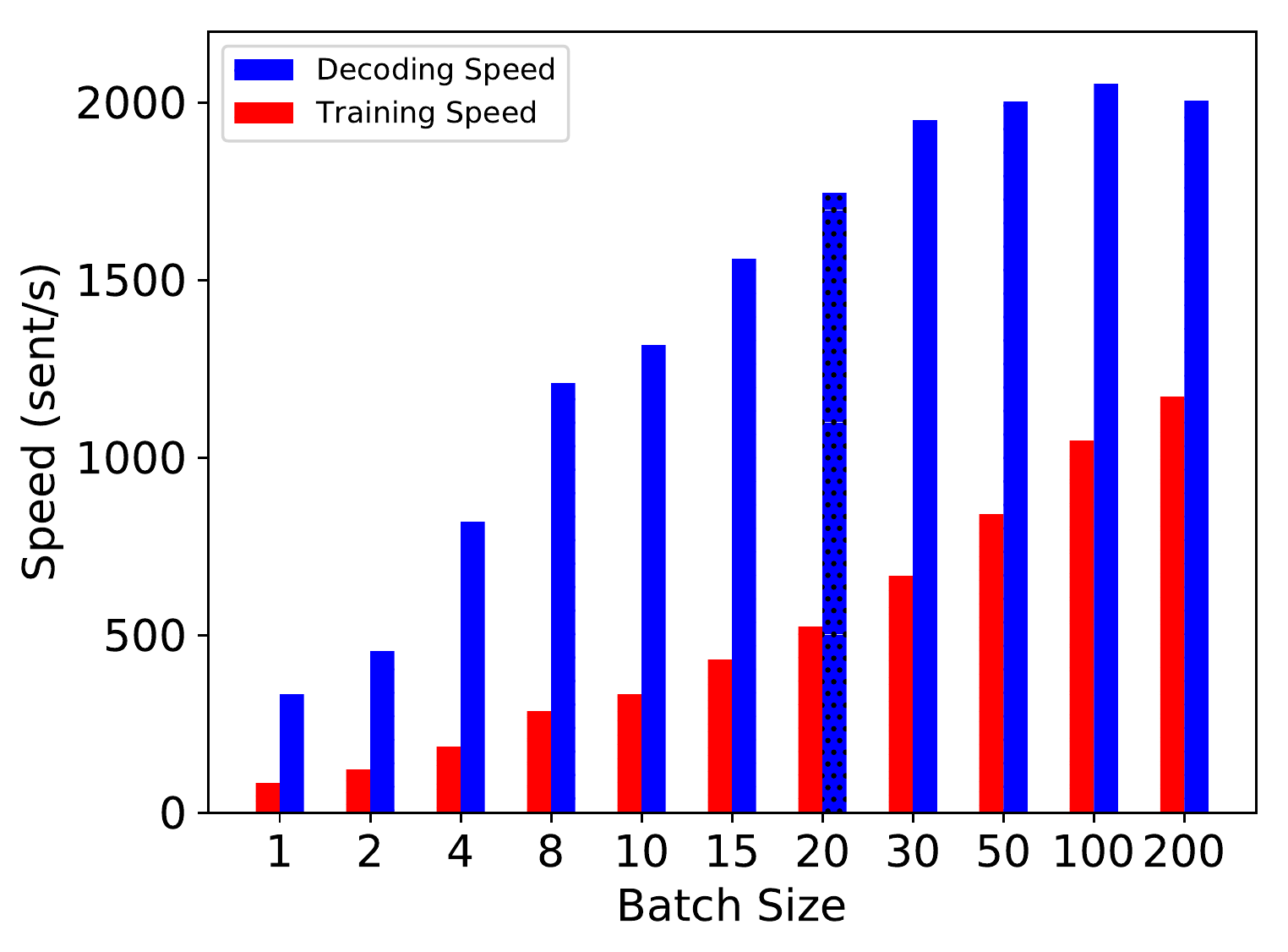}
  \caption{Speed with batch size.} 
  \label{fig:speed} 
\end{figure}

As NCRF++ is implemented on batched calculation, it can be greatly accelerated through parallel computing through GPU. We test the system speeds on both training and decoding process on NER dataset using a Nvidia GTX 1080 GPU. As shown in Figure \ref{fig:speed}, both the training and the decoding speed can be significantly accelerated through a large batch size. The decoding speed reaches saturation at batch size 100, while the training speed keeps growing. The decoding speed and training speed of NCRF++ are over 2000 sentences/second and 1000 sentences/second, respectively, demonstrating the efficiency of our implementation.

\section{Conclusion}
We presented NCRF++, an open-source neural sequence labeling toolkit, which has a CRF architecture with configurable neural representation layers. Users can design custom neural models through the configuration file. NCRF++ supports flexible feature utilization, including handcrafted features and automatically extracted features. It can also generate \textit{nbest} label sequences rather than the best one. We conduct a series of experiments and the results show models built on NCRF++ can achieve state-of-the-art results with an efficient running speed.


\newpage
\bibliography{acl2018}

\begin{thebibliography}{}
\expandafter\ifx\csname natexlab\endcsname\relax\def\natexlab#1{#1}\fi

\bibitem[{Chiu and Nichols(2016)}]{chiu2015named}
Jason Chiu and Eric Nichols. 2016.
\newblock \href{https://transacl.org/ojs/index.php/tacl/article/view/792}{Named
  entity recognition with bidirectional {LSTM-CNNs}}.
\newblock {\em Transactions of the Association for Computational Linguistics\/}
  4:357--370.
\newblock
  \href{https://transacl.org/ojs/index.php/tacl/article/view/792}{https://transacl.org/ojs/index.php/tacl/article/view/792}.

\bibitem[{Collobert et~al.(2011)Collobert, Weston, Bottou, Karlen, Kavukcuoglu,
  and Kuksa}]{collobert2011natural}
Ronan Collobert, Jason Weston, L{\'e}on Bottou, Michael Karlen, Koray
  Kavukcuoglu, and Pavel Kuksa. 2011.
\newblock Natural language processing (almost) from scratch.
\newblock {\em Journal of Machine Learning Research\/} 12(Aug):2493--2537.

\bibitem[{Glorot et~al.(2011)Glorot, Bordes, and Bengio}]{glorot2011deep}
Xavier Glorot, Antoine Bordes, and Yoshua Bengio. 2011.
\newblock Deep sparse rectifier neural networks.
\newblock In {\em International Conference on Artificial Intelligence and
  Statistics\/}. pages 315--323.

\bibitem[{Hochreiter and Schmidhuber(1997)}]{hochreiter1997long}
Sepp Hochreiter and J{\"u}rgen Schmidhuber. 1997.
\newblock Long short-term memory.
\newblock {\em Neural computation\/} 9(8):1735--1780.

\bibitem[{Huang et~al.(2015)Huang, Xu, and Yu}]{huang2015bidirectional}
Zhiheng Huang, Wei Xu, and Kai Yu. 2015.
\newblock Bidirectional {LSTM-CRF} models for sequence tagging.
\newblock {\em arXiv preprint arXiv:1508.01991\/} .

\bibitem[{Ioffe and Szegedy(2015)}]{ioffe2015batch}
Sergey Ioffe and Christian Szegedy. 2015.
\newblock Batch normalization: Accelerating deep network training by reducing
  internal covariate shift.
\newblock In {\em International Conference on Machine Learning\/}. pages
  448--456.

\bibitem[{Keerthi and Sundararajan(2007)}]{keerthi2007crf}
S~Sathiya Keerthi and Sellamanickam Sundararajan. 2007.
\newblock Crf versus svm-struct for sequence labeling.
\newblock {\em Yahoo Research Technical Report\/} .

\bibitem[{Lafferty et~al.(2001)Lafferty, McCallum, and
  Pereira}]{lafferty2001conditional}
John Lafferty, Andrew McCallum, and Fernando~CN Pereira. 2001.
\newblock Conditional random fields: Probabilistic models for segmenting and
  labeling sequence data.
\newblock In {\em International Conference on Machine Learning\/}. volume~1,
  pages 282--289.

\bibitem[{Lample et~al.(2016)Lample, Ballesteros, Subramanian, Kawakami, and
  Dyer}]{lample2016neural}
Guillaume Lample, Miguel Ballesteros, Sandeep Subramanian, Kazuya Kawakami, and
  Chris Dyer. 2016.
\newblock Neural architectures for named entity recognition.
\newblock In {\em NAACL-HLT\/}. pages 260--270.

\bibitem[{LeCun et~al.(1989)LeCun, Boser, Denker, Henderson, Howard, Hubbard,
  and Jackel}]{lecun1989backpropagation}
Yann LeCun, Bernhard Boser, John~S Denker, Donnie Henderson, Richard~E Howard,
  Wayne Hubbard, and Lawrence~D Jackel. 1989.
\newblock Backpropagation applied to handwritten zip code recognition.
\newblock {\em Neural computation\/} 1(4):541--551.

\bibitem[{Ling et~al.(2015)Ling, Dyer, Black, Trancoso, Fermandez, Amir,
  Marujo, and Luis}]{ling2015finding}
Wang Ling, Chris Dyer, Alan~W Black, Isabel Trancoso, Ramon Fermandez, Silvio
  Amir, Luis Marujo, and Tiago Luis. 2015.
\newblock Finding function in form: Compositional character models for open
  vocabulary word representation.
\newblock In {\em EMNLP\/}. pages 1520--1530.

\bibitem[{Liu et~al.(2018)Liu, Shang, Xu, Ren, Gui, Peng, and
  Han}]{liu2017empower}
Liyuan Liu, Jingbo Shang, Frank Xu, Xiang Ren, Huan Gui, Jian Peng, and Jiawei
  Han. 2018.
\newblock Empower sequence labeling with task-aware neural language model.
\newblock In {\em AAAI\/}.

\bibitem[{Ma and Hovy(2016)}]{ma2016end}
Xuezhe Ma and Eduard Hovy. 2016.
\newblock End-to-end sequence labeling via {B}i-directional {LSTM-CNNs-CRF}.
\newblock In {\em ACL\/}. volume~1, pages 1064--1074.

\bibitem[{Okazaki(2007)}]{okazaki2007crfsuite}
Naoaki Okazaki. 2007.
\newblock Crfsuite: a fast implementation of conditional random fields (crfs) .

\bibitem[{Peters et~al.(2017)Peters, Ammar, Bhagavatula, and
  Power}]{peters2017semi}
Matthew Peters, Waleed Ammar, Chandra Bhagavatula, and Russell Power. 2017.
\newblock Semi-supervised sequence tagging with bidirectional language models.
\newblock In {\em ACL\/}. volume~1, pages 1756--1765.

\bibitem[{Phan et~al.(2004)Phan, Nguyen, and Nguyen}]{phan2004flexcrfs}
Xuan-Hieu Phan, Le-Minh Nguyen, and Cam-Tu Nguyen. 2004.
\newblock Flexcrfs: Flexible conditional random fields.

\bibitem[{Ratinov and Roth(2009)}]{ratinov2009design}
Lev Ratinov and Dan Roth. 2009.
\newblock Design challenges and misconceptions in named entity recognition.
\newblock In {\em CoNLL\/}. pages 147--155.

\bibitem[{Reimers and Gurevych(2017)}]{reimers2017reporting}
Nils Reimers and Iryna Gurevych. 2017.
\newblock Reporting score distributions makes a difference: Performance study
  of lstm-networks for sequence tagging.
\newblock In {\em EMNLP\/}. pages 338--348.

\bibitem[{Sha and Pereira(2003)}]{sha2003shallow}
Fei Sha and Fernando Pereira. 2003.
\newblock Shallow parsing with conditional random fields.
\newblock In {\em NAACL-HLT\/}. pages 134--141.

\bibitem[{Srivastava et~al.(2014)Srivastava, Hinton, Krizhevsky, Sutskever, and
  Salakhutdinov}]{srivastava2014dropout}
Nitish Srivastava, Geoffrey~E Hinton, Alex Krizhevsky, Ilya Sutskever, and
  Ruslan Salakhutdinov. 2014.
\newblock Dropout: a simple way to prevent neural networks from overfitting.
\newblock {\em Journal of Machine Learning Research\/} 15(1):1929--1958.

\bibitem[{Strubell et~al.(2017)Strubell, Verga, Belanger, and
  McCallum}]{strubell2017fast}
Emma Strubell, Patrick Verga, David Belanger, and Andrew McCallum. 2017.
\newblock Fast and accurate entity recognition with iterated dilated
  convolutions.
\newblock In {\em EMNLP\/}. pages 2670--2680.

\bibitem[{Tjong Kim~Sang and Buchholz(2000)}]{tjong2000introduction}
Erik~F Tjong Kim~Sang and Sabine Buchholz. 2000.
\newblock Introduction to the conll-2000 shared task: Chunking.
\newblock In {\em Proceedings of the 2nd workshop on Learning language in logic
  and the 4th conference on Computational natural language learning-Volume
  7\/}. pages 127--132.

\bibitem[{Tjong Kim~Sang and De~Meulder(2003)}]{tjong2003introduction}
Erik~F Tjong Kim~Sang and Fien De~Meulder. 2003.
\newblock Introduction to the conll-2003 shared task: Language-independent
  named entity recognition.
\newblock In {\em HLT-NAACL\/}. pages 142--147.

\bibitem[{Viterbi(1967)}]{viterbi1967error}
Andrew Viterbi. 1967.
\newblock Error bounds for convolutional codes and an asymptotically optimum
  decoding algorithm.
\newblock {\em IEEE transactions on Information Theory\/} 13(2):260--269.

\bibitem[{Yang et~al.(2018)Yang, Liang, and Zhang}]{yang2018design}
Jie Yang, Shuailong Liang, and Yue Zhang. 2018.
\newblock Design challenges and misconceptions in neural sequence labeling.
\newblock In {\em COLING\/}.

\bibitem[{Yang et~al.(2017)Yang, Salakhutdinov, and Cohen}]{yang2017transfer}
Zhilin Yang, Ruslan Salakhutdinov, and William~W Cohen. 2017.
\newblock Transfer learning for sequence tagging with hierarchical recurrent
  networks.
\newblock In {\em International Conference on Learning Representations\/}.

\end{thebibliography}
\bibliographystyle{acl_natbib}

\end{document}